\documentclass[10pt,twocolumn,letterpaper]{article}
\pdfoutput=1
\usepackage{cvpr}
\usepackage{times}
\usepackage{epsfig}
\usepackage{graphicx}
\usepackage{amsmath}
\usepackage{amssymb}
\usepackage{xcolor}
\usepackage{multicol}
\usepackage{gensymb}
\usepackage{booktabs}
\usepackage{multirow}
\usepackage{subcaption}

\usepackage[breaklinks=true,bookmarks=false]{hyperref}

\cvprfinalcopy 

\def\cvprPaperID{6119} 

\title{LaserNet: An Efficient Probabilistic 3D Object Detector for Autonomous Driving}

\author{Gregory P. Meyer*, Ankit Laddha*, Eric Kee, Carlos Vallespi-Gonzalez, Carl K. Wellington\\
Uber Advanced Technologies Group\\
{\{\tt\small gmeyer,aladdha,ekee,cvallespi,cwellington\}@uber.com}}

\ifcvprfinal\fi
\begin{document}

\maketitle
\thispagestyle{empty}
\makeatletter{\renewcommand*{\@makefnmark}{}
\footnotetext{*Equal Contribution}\makeatother}


%

\begin{abstract}

In  this  paper,  we  present  LaserNet,  a  computationally efficient method for 3D object detection from LiDAR data for autonomous driving. The efficiency results from processing LiDAR data in the native range view of the sensor, where the input data is naturally compact. Operating in the range view involves well known challenges for learning, including occlusion and scale variation, but it also provides contextual information based on how the sensor data was captured. Our approach uses a fully convolutional network to predict a multimodal distribution over 3D boxes for each point and then it efficiently fuses these distributions to generate a prediction for each object. Experiments show that modeling each detection as a distribution rather than a single deterministic box leads to better overall detection performance. Benchmark results show that this approach has significantly lower runtime than other recent detectors and that it achieves state-of-the-art performance when compared on a large dataset that has enough data to overcome the challenges of training on the range view.


\end{abstract}

\section{Introduction}


3D object detection is a key capability for autonomous driving. LiDAR range sensors are commonly used for this task because they generate accurate range measurements of the objects of interest independent of lighting conditions. To be used in a real-time autonomous system, it is important that these approaches run efficiently in addition to having high accuracy. Also, within the context of a full self-driving system, it is beneficial to have an understanding of the detector's uncertainty.

\begin{figure}[t]
\centering
\includegraphics[width=0.9\columnwidth]{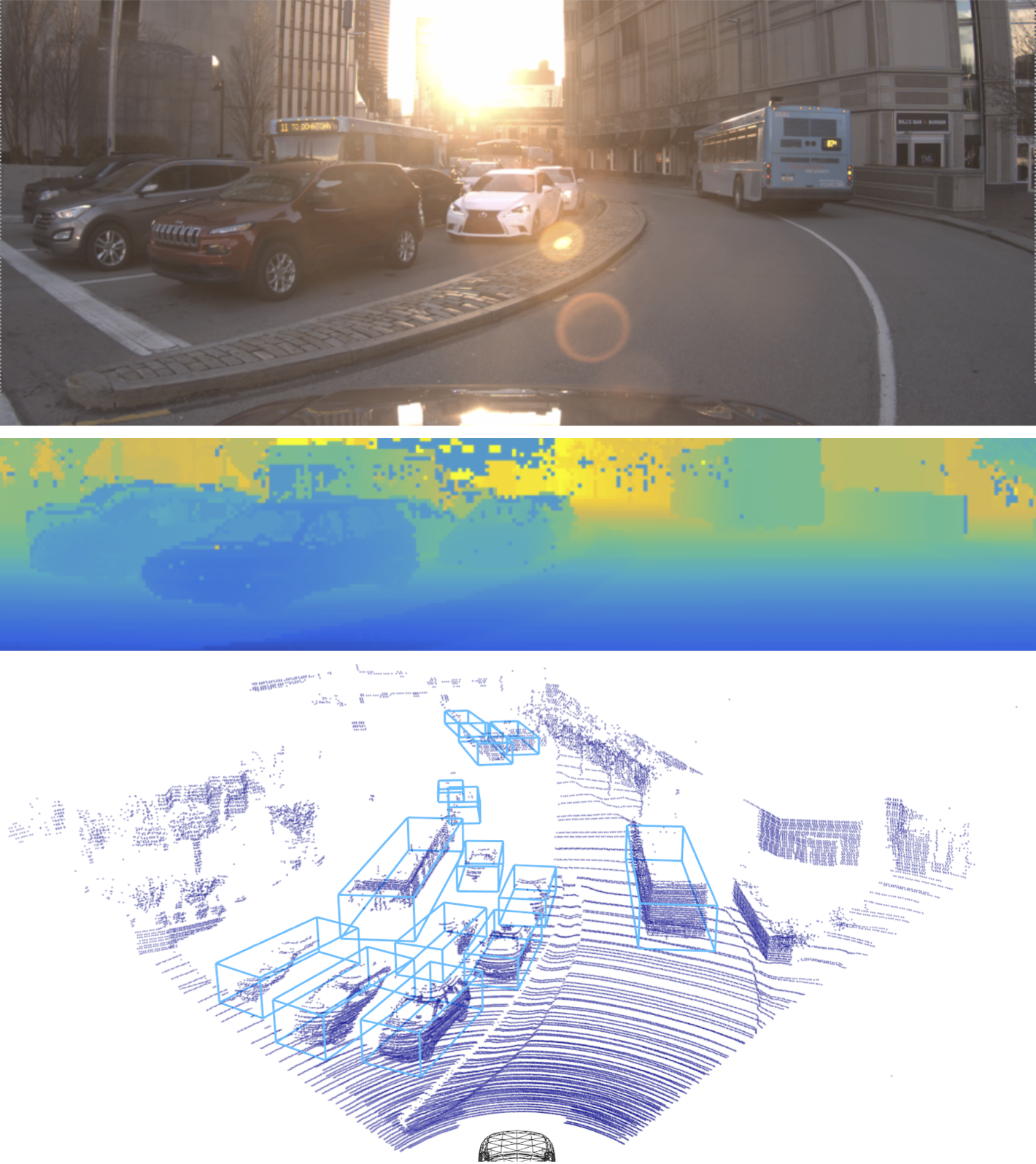}
\caption{Example detection results from our approach (bottom) generated from the input range image (middle). The camera image (top) is provided for reference. Notice the data is dense in the range image but becomes sparse when projected into 3D space.}
\label{fig:fv_bev}
\end{figure}

LiDAR range sensors used on autonomous vehicles employ a variety of physical approaches to point one or more range-measuring lasers in different directions (physically spinning, raster scanning, MEMS mirrors, etc), but the data from these sensors are all similar in that they contain range measurements from a single location spanning some vertical and horizontal angular field of view. Note that this data is quite different from full 3D point cloud data collected by scanning an object on a turntable or through other means to generate a consistent dense point cloud of all sides of the object. As shown in Figure~\ref{fig:fv_bev}, the LiDAR data is inherently dense from the sensor's point of view but sparse when the range measurements are projected into 3D space. The density of measurements is constant in angle, so nearby objects have significantly more measurements than far away objects. Also, only the sides of objects facing the sensor receive measurements.

When performing 2D detection in camera images, efficient and high-performing results have been achieved using dense convolutional approaches~\cite{ssd, faster_rcnn}.
Although the sensor's range data is similar in many ways to a camera image, the desired output is an oriented bounding box in 3D space. Therefore, 3D object detectors must transform the sensor data in the range view (RV) into detections in the top-down view also known as the bird's eye view (BEV). Most existing work starts by projecting the range measurements into 3D points and then discretizing the 3D space into voxels in order to operate directly in the output space~\cite{hdnet, pixor, ibm}. These approaches have shown impressive results but require operating on sparse input data which can be computationally expensive. There has also been work operating on the dense RV representation directly~\cite{mv3d, velofcn}, but these methods have not matched the performance of the BEV approaches. Additionally, there has been work combining both RV and BEV representations~\cite{mv3d}.

These representations each have different advantages and disadvantages.
In the RV, the sensor data is dense, but the perceived size of objects varies with range.
In the BEV, the data is sparse, but the size of objects remains constant regardless of range.
This consistency adds a strong prior to the predictions making the problem easier to learn.
Lastly, the RV preserves occlusion information which is lost when projecting the data into the BEV.

In this work, we present an efficient method for learning a probabilistic 3D object detector in an end-to-end fashion. When there is sufficient training data, we achieve state-of-the-art detection performance with a significantly lower runtime. Our approach is efficient because we use a small and dense range image instead of a large and sparse bird's eye view image. Our proposed method produces not only a class probability for each detection but also a probability distribution over detection bounding boxes. To our knowledge, our proposed method is the first to capture the uncertainty of a detection by modeling the distribution of bounding box corners. By estimating the accuracy of a detection, our approach enables downstream components in a full self-driving system to behave differently around objects with varying levels of uncertainty.

\begin{figure*}[t]
\centering
\includegraphics[width=\textwidth]{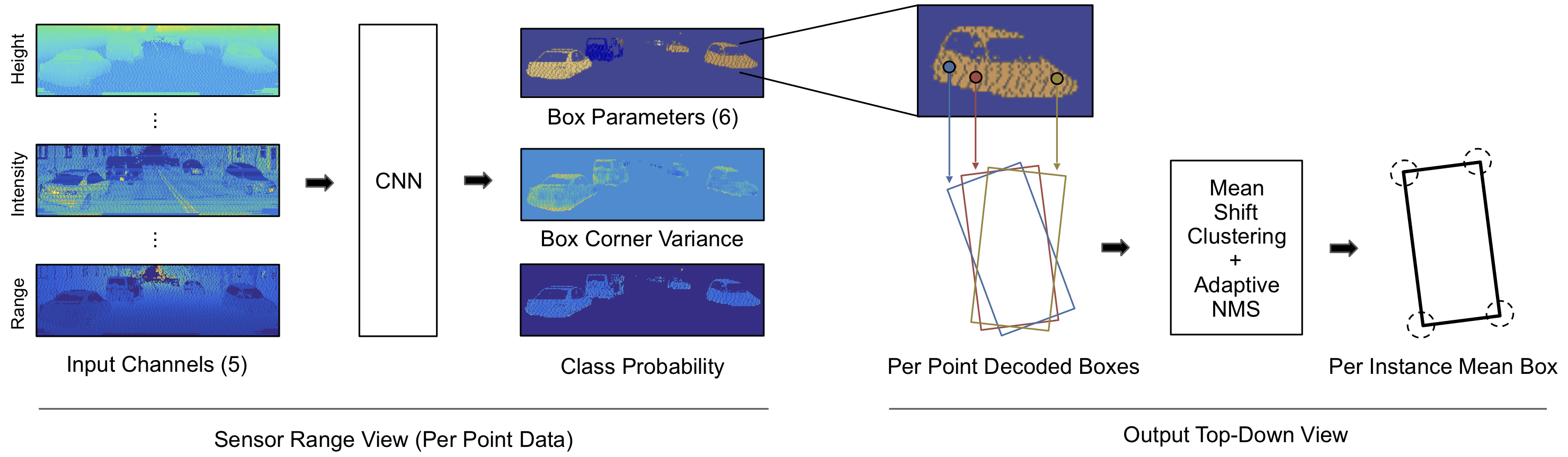}
\caption{An overview of our approach to 3D object detection. We use the inherent range view representation of the sensor to build a dense input image (Section~\ref{sec:input}). The image is passed through a fully-convolutional network to produce a set of predictions (Section~\ref{sec:network}). For each LiDAR point in the image, we predict a class probability, and we regress a probability distribution over bounding boxes in the top-down view (Section~\ref{sec:predictions}). These per-point distributions are combined through mean shift clustering to reduce the noise in the individual predictions (Section~\ref{sec:clustering}). The entire detector is trained end-to-end with the loss defined on the box corners (Section~\ref{sec:training}). At inference, we leverage a novel adaptive non-maximum suppression (NMS) algorithm to remove duplicate box distributions (Section~\ref{sec:nms}).}
\label{fig:overview}
\end{figure*}

\section{Related Work}


\subsection{3D Object Detection}
Multiple deep learning based methods have been proposed for 3D object detection for autonomous driving. VeloFCN \cite{velofcn} and MV3D \cite{mv3d} proposed methods using a range view representation. Both methods construct the RV by discretizing the azimuth and elevation angles. In this paper, we propose a different RV representation using the configuration of the LiDAR, and demonstrate that it leads to better performance.

VoxelNet \cite{voxelnet} computes PointNet \cite{pointnet} style features from the LiDAR points for each 3D voxel. Afterwards, they run 3D convolutions over these voxels to produce detections. Instead of voxelizing the input, our approach operates directly on the raw range data and aggregates predictions from LiDAR points through mean shift clustering to generate the detections.

Similar to 2D image detection, 3D methods can be grouped into two meta frameworks: region proposal methods \cite{faster_rcnn} and single shot methods \cite{ssd}. Region proposal based methods \cite{mv3d, avod, frustum_pointnet} have two stages: the first stage proposes plausible regions containing objects, and the second stage extracts features from the proposed regions and uses them to produce detections. Single shot based methods \cite{birdnet, velofcn, pixor, voxelnet} produce detections with a single stage network. In this work, we use a single shot approach since region proposal networks are often computationally expensive and minimizing latency is critical for a real-time autonomous system.

In addition to using LiDAR data, several previous works use images \cite{mv3d, avod, frustum_pointnet, ibm, pointfusion, rgb_dense_lidar_fusion, rgb_dense_lidar_multiview_fusion} or high definition maps \cite{hdnet} to improve detections. In this paper, we only utilize LiDAR data and leave sensor fusion as future work. Some approaches \cite{nips15chen, cvpr16chen, cvpr17mousavian, cvpr18xu} have tried to tackle 3D detection without using range data from LiDAR. However, the accuracy of these methods do not reach the performance of LiDAR based methods.

\subsection{Probabilistic Object Detection}
Most of the state-of-the-art 2D \cite{faster_rcnn, ssd} and 3D \cite{mv3d, avod, hdnet, ibm} object detection methods produce a single box with a probability score for each detection. While this probability encompasses the existence and semantic uncertainty, it does not accurately measure the localization uncertainty. Recently, \cite{uncertainty_2d_detection} proposed to explicitly predict the intersection-over-union (IoU) of each 2D detection with the ground truth label. They use the predicted IoU to refine and score the detections. Estimating the IoU provides a measure of the detection's localization error. In this paper, we estimate a probability distribution over box detections instead of only the mean, and the variance of the predicted distribution indicates the amount of uncertainty in the position of the box corners.

The seminal work in \cite{kendalluncertainities} studies Bayesian deep networks for multiple computer vision tasks.
Based on this work, \cite{uncertainty_3d_detection_itsc, uncertainty_3d_detection} produce and analyze epistemic and aleatoric uncertainties in 3D object detection.
However, they do not use the predicted uncertainties to significantly improve the performance of the detector.
In this paper, we focus on estimating the aleatoric uncertainty by way of predicting a probability distribution over box detections, and we leverage the predicted distribution to combine and refine our detections.
The epistemic uncertainty is not estimated by our method since it currently cannot be computed efficiently.

\section{Proposed Method}
An overview of our proposed method is shown in Figure \ref{fig:overview}.
In the following sections each step is described in detail.

\subsection{Input Representation}
\label{sec:input}
The LiDAR produces a cylindrical range image as it sweeps over the environment with a set of lasers. The horizontal resolution of the image is determined by the rotation speed and laser pulse rate, and the vertical resolution is determined by the number of lasers. The Velodyne 64E LiDAR contains a set of $64$ lasers with a non-uniform vertical spacing of approximately $0.4^\circ$ and has a horizontal angular resolution of approximately $0.2^\circ$.
For each point in the sweep, the sensor provides a range $r$, reflectance $e$, azimuth $\theta$, and laser id $m$, which corresponds to a known elevation angle. Using the range, azimuth, and elevation, we can compute the corresponding 3D point $(x,y,z)$ in the sensor frame. We build an input image by directly mapping the laser id $m$ to rows and discretizing azimuth $\theta$ into columns. If more than one point occupies the same cell in the image, we keep the closest point.

For each cell coordinate in the image, we collect a set of input channels from its corresponding point: range $r$, height $z$, azimuth angle $\theta$, intensity $e$, and a flag indicating whether the cell contains a point. The result is a five channel image that forms the input to our network (see Figure~\ref{fig:overview}).

\subsection{Network Architecture}
\label{sec:network}
\begin{figure}
\centering
\includegraphics[width=0.4\textwidth]{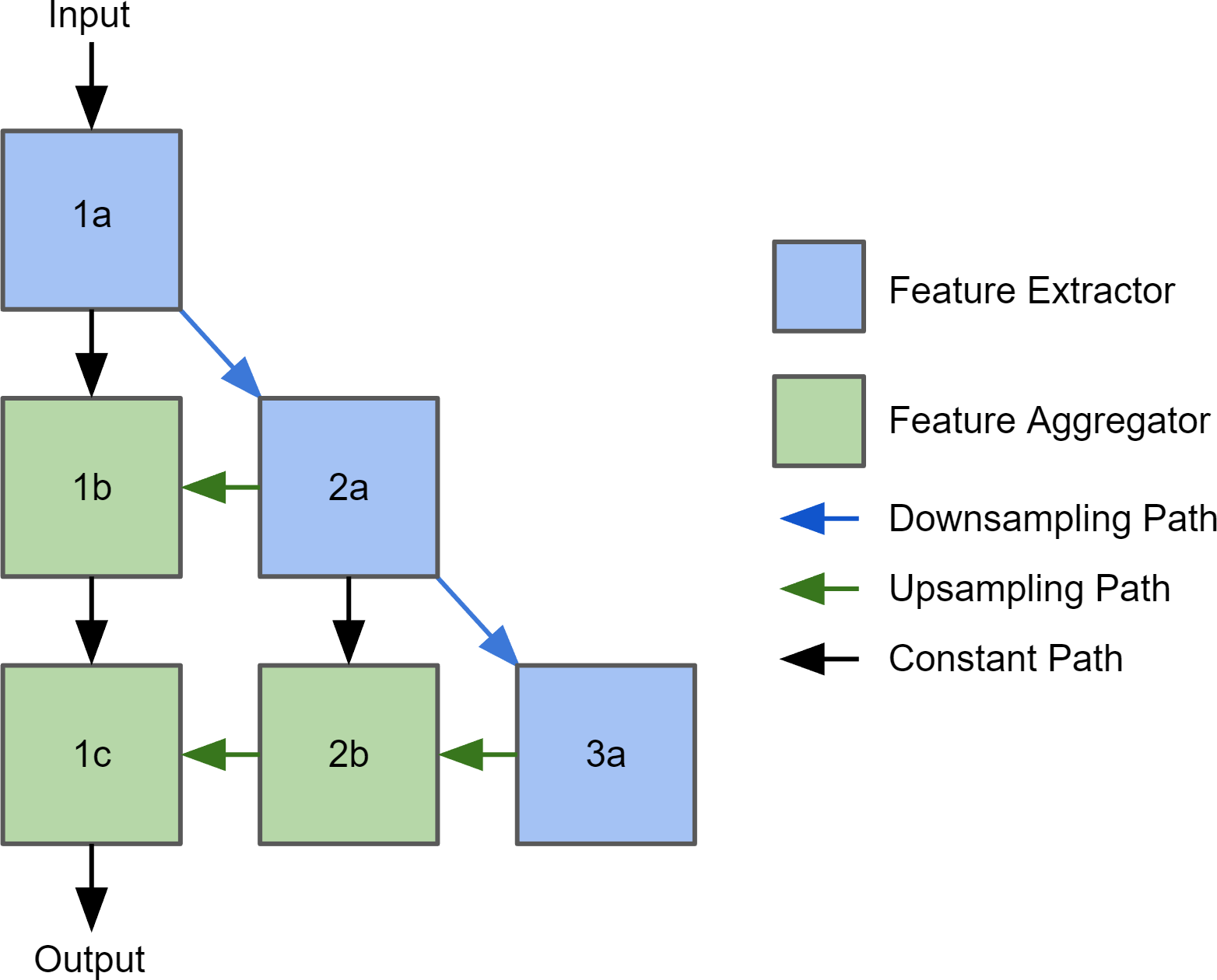}
\caption{Our deep layer aggregation network architecture. The columns indicate different resolution levels, and the rows indicate aggregation stages.}
\label{fig:network}
\end{figure}

Our image contains objects at a wide range of distances, and the size of an object can vary from several thousand points to a single point.
We leverage the deep layer aggregation network architecture \cite{dla} to effectively extract and combine multi-scale features.
Our network is fully-convolutional and consists of three hierarchical levels as shown in Figure \ref{fig:network}.
The size of convolutional kernels at each level is 64, 64, 128 respectively.

\begin{figure}
\centering
\includegraphics[width=0.35\textwidth]{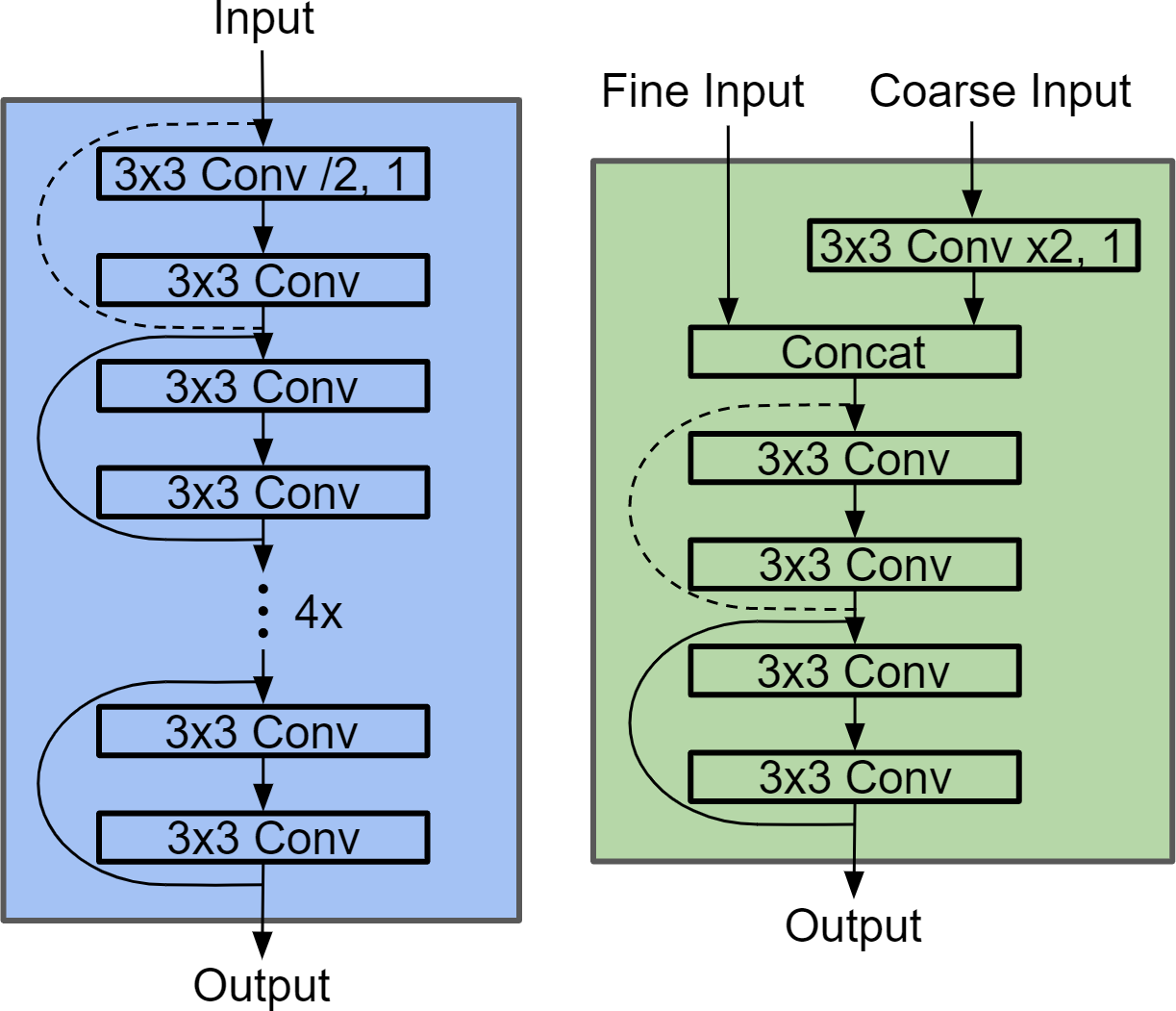}
\caption{Our feature extraction module (left) and feature aggregation module (right). The modules are residual networks \cite{resnet}. Dashed lines indicate a convolution is performed to reshape the feature map.}
\label{fig:network_modules}
\end{figure}

Each level contains a feature extraction module and some number of feature aggregation modules.
The structure of these modules is depicted in Figure \ref{fig:network_modules}.
Since the horizontal resolution of the image is significantly larger than the vertical resolution, we keep the vertical resolution constant and only perform downsampling and upsampling along the horizontal dimension.
A final $1\times1$ convolutional layer is used to transform the resulting feature map to our encoded predictions.

\subsection{Predictions}
\label{sec:predictions}
Our network is trained to predict a set of class probabilities for each point in the image.
Given a point lies on an object, the network predicts a probability distribution over bounding boxes that could encapsulate the object.
The probability distribution can be multimodal when the observed data is incomplete, which often occurs at long range, when points are sparse, and when there are occlusions.
Therefore, we model the probability distribution with a mixture model, and train the network to predict a set of means with corresponding variances and mixture weights.
The distribution learned by the network is defined by the loss function, which is discussed in Section~\ref{sec:training}.

For the application of autonomous driving, we assume all objects lie on the same ground plane; therefore, we can characterize a bounding box by its four corners in the \mbox{$x$-$y$} plane.
The 3D bounding boxes depicted in Figure~\ref{fig:fv_bev} are produced using an assumed ground plane and a fixed height.
Instead of regressing the four corners directly, we predict a relative center $\left(d_x, d_y \right)$, a relative orientation $\left(\omega_x, \omega_y \right) = \left(\cos\omega, \sin\omega \right)$, and the dimensions $\left(l, w \right)$ of the box.
We compute the box's absolute center $\boldsymbol{b}_c$ and absolute orientation $\phi$ as follows:
\begin{equation} \label{eqn:box-center}
\begin{aligned}
\boldsymbol{b}_c &= \left[ x, y \right]^T + \boldsymbol{R}_\theta \left[ d_x, d_y \right]^T \\
\phi &= \theta + \text{atan2}\left(\omega_y, \omega_x \right)
\end{aligned}
\end{equation}
where $\left(x, y \right)$ and $\theta$ is the 2D position and azimuth angle of the LiDAR point, and $\boldsymbol{R}_\theta$ is the rotation matrix parameterized by $\theta$.
Subsequently, we can calculate the four corners of the bounding box,
\begin{equation} \label{eqn:box-corners}
\begin{aligned}
\boldsymbol{b}_1 &= \boldsymbol{b}_c + \frac{1}{2} \boldsymbol{R}_\phi \left[ l, w \right]^T \\
\boldsymbol{b}_3 &= \boldsymbol{b}_c + \frac{1}{2} \boldsymbol{R}_\phi \left[ -l, -w \right]^T
\end{aligned}
\quad
\begin{aligned}
\boldsymbol{b}_2 &= \boldsymbol{b}_c + \frac{1}{2} \boldsymbol{R}_\phi \left[ l, -w \right]^T \\
\boldsymbol{b}_4 &= \boldsymbol{b}_c + \frac{1}{2} \boldsymbol{R}_\phi \left[ -l, w \right]^T
\end{aligned}
\end{equation}
where $\boldsymbol{R}_\phi$ is the rotation matrix parameterized by $\phi$.
For convenience, we concatenate the four corners into a single vector, $\boldsymbol{b} = [\boldsymbol{b}_1,  \boldsymbol{b}_2,  \boldsymbol{b}_3, \boldsymbol{b}_4$].

To simplify the predict probability distribution, we assume a uniform variance across both the $x$ and $y$ dimensions, and we share the variance across all four corners of the bounding box.
As in \cite{kendalluncertainities}, we train the network to predict the log of the standard deviation $s = \log \sigma$.

Altogether, for each point in the image, our network predicts a set of class probabilities $\left\{ p_c \right\}_{c=1}^{C}$, where $C$ is the number of object classes in addition to a background class.
For each point and class of object, the network predicts a set of bounding box parameters $\left\{ d_{x,k}, d_{y,k}, \omega_{x,k}, \omega_{y,k}, l_k, w_k \right\}_{k=1}^K$, a set of log standard deviations $\left\{ s_k \right\}_{k=1}^K$, and a set of mixture weights $\left\{ \alpha_k \right\}_{k=1}^K$, where $K$ is the number of components in the mixture model.

\subsection{Mean Shift Clustering}
\label{sec:clustering}
Each point independently predicts the distribution of bounding boxes; however, points on the same object should predict a similar distribution.
Naturally, the individual predictions will contain some amount of noise.
We can reduce this noise by combining the per-point predictions through mean shift clustering.
Since our predicted distribution is class dependent and multimodal, we perform mean shift on each object class and each component of the mixture model separately.
For efficiency, mean shift is performed over box centers instead of box corners, which reduces the dimensionality of the problem.
Additionally, we propose a fast approximation of the mean shift algorithm when operating in two dimensions.

Our approximate algorithm begins by discretizing the top-down view into bins of size $\Delta x$ by $\Delta y$.
For each bin $i$, which contains one or more box centers, we create an initial mean $\boldsymbol{m}_i$ by averaging over all the centers in that bin.
In addition, we record the set of points $S_i$ whose box centers fall inside the bin.
We iteratively update the means as follows:
\begin{equation} \label{eqn:mean-shift}
\boldsymbol{m}_i \leftarrow \frac{\sum_{j \in i \cup N(i)} K_{i,j} \left( \boldsymbol{m}_j \cdot |S_j| \right)}{\sum_{j \in i \cup N(i)} K_{i,j} |S_j|}
\end{equation}
where
\begin{equation}
K_{i,j} = \exp\left(-\frac{\|\boldsymbol{m}_i - \boldsymbol{m}_j\|^2}{\Delta x^2 + \Delta y^2}\right)
\end{equation}
and $N(i)$ is the set of eight bins neighboring the $i$th bin.
After the update, if the $i$th mean now falls into the $j$th bin, we merge $i$ into $j$,
\begin{equation}
\begin{aligned}
\boldsymbol{m}_j \leftarrow \frac{\boldsymbol{m}_i \cdot |S_i| + \boldsymbol{m}_j \cdot |S_j|}{|S_i| + |S_j|} &&
S_j \leftarrow S_i \cup S_j
\end{aligned}
\end{equation}
and invalidate $\boldsymbol{m}_i \leftarrow \boldsymbol{0}$ and $S_i \leftarrow \emptyset$.

Equation (\ref{eqn:mean-shift}) can be computed efficiently by exploiting the regular structure of the bins.
By constructing a tensor of bins and generating shifted versions of this tensor, we can update all the means simultaneously using only element-wise operators.
This type of computation is well suited for a graphics processing unit (GPU).

After performing a fixed number of mean shift iterations, per-point box distributions that are assigned to the same cluster are combined.
Specifically, we model the cluster bounding box distribution as the product of the per-point distributions.
The mean and variance of the cluster probability distribution is
\begin{equation} \label{eqn:cluster-corners}
\begin{aligned}
\hat{\boldsymbol{b}}_{i} = \frac{\sum_{j \in S_i} w_j \boldsymbol{b}_{j}}{\sum_{j \in S_i} w_j} &&
\hat{\sigma}_i^2 = \bigg(\sum_{j \in S_i} \frac{1}{\sigma_j^2} \bigg)^{-1}
\end{aligned}
\end{equation}
where $w = 1/\sigma^2$. Each point's predicted bounding box and variance is replaced by the bounding box and variance of its cluster.
For our experiments, we perform three iterations of mean shift with $\Delta x = \Delta y = 0.5$ meters.

\subsection{End-to-end Training} \label{sec:training}
Mapping from box parameters to box corners, equations (\ref{eqn:box-center}) and (\ref{eqn:box-corners}), and merging the bounding boxes distributions, equation (\ref{eqn:cluster-corners}), are differentiable; therefore, we are able to train the network in an end-to-end fashion.

For each point in the image, we use the multi-class cross entropy loss $\mathcal{L}_{\text{prob}}$ to learn the class probabilities $\left\{ p_c \right\}_{c=1}^{C}$.
To handle class imbalance, we employ focal loss \cite{retinanet}, which is a modified version of the cross entropy loss.
The classification loss for the entire image is defined as follows:
\begin{equation}
\mathcal{L}_{\text{cls}} = \frac{1}{P} \sum_i \mathcal{L}_{\text{prob}, i}
\end{equation}
where $\mathcal{L}_{\text{prob}, i}$ is the loss for the $i$th point in the image, and $P$ is the total number of points in the image.

For each point on an object, we learn the parameters of the object's mixture model by first identifying which component best matches the ground truth,
\begin{equation}
k^* = \arg\min_k \| \hat{\boldsymbol{b}}_{k} - \boldsymbol{b}^\text{gt} \|
\end{equation}
where $\hat{\boldsymbol{b}}_{k}$ is the $k$th mean component of the mixture model and $\boldsymbol{b}^\text{gt}$ is the corresponding ground truth bounding box.
Afterwards, we update the parameters of the $k^*$ component following the approach proposed in \cite{kendalluncertainities},
\begin{equation} \label{eqn:cluster-loss}
\mathcal{L}_{\text{box}} = \sum_{n} \frac{1}{\hat{\sigma}_{k^*}} \left|\hat{b}_{n, k^*} - b_n^\text{gt} \right|   +  \log \hat{\sigma}_{k^*}
\end{equation}
where $\hat{b}_{n, k^*}$ is the $n$th element of $\hat{\boldsymbol{b}}_{k^*}$, and $b_n^\text{gt}$ is the corresponding ground truth value.
As discussed in \cite{kendalluncertainities}, this loss imposes a Laplacian prior on the learned distribution.
Next, we update the mixture weights $\left\{ \alpha_k \right\}_{k=1}^K$ again using the multi-class cross entropy loss $\mathcal{L}_{\text{mix}}$, where the positive label corresponds to the $k^*$ component.
The idea of only updating the prediction that best matches the ground truth was originally proposed in \cite{multiple-choice} as the hindsight loss for multiple choice learning.
The regression loss for the entire image is defined as follows:
\begin{equation}
\mathcal{L}_{\text{reg}} = \frac{1}{N} \sum_i \frac{\mathcal{L}_{\text{box}, i} + \lambda \mathcal{L}_{\text{mix}, i}}{n_i}
\end{equation}
where $\mathcal{L}_{\text{box}, i}$ and $\mathcal{L}_{\text{mix}, i}$ are the losses for the $i$th point in the image which is on an object, $n_i$ is the total number of points that lie on the same object as $i$, $N$ is the total instances of objects in the image, and $\lambda$ is the relative weighting of the two losses. We set $\lambda=0.25$ for all our experiments. The total loss for the image is $\mathcal{L}_{\text{total}} = \mathcal{L}_{\text{cls}} + \mathcal{L}_{\text{reg}}$.

\subsection{Adaptive Non-Maximum Suppression}
\label{sec:nms}
At inference, we identify the points that belong to an object class by thresholding the predicted class probability $p_{c}$, and we use a threshold of $1 / C$ in all our experiments.
As described previously, each point on an object predicts a probability distribution over bounding boxes.
For each predicted distribution, we extract a set of bounding boxes which correspond to the means of the mixture model.

Typically non-maximum suppression (NMS) is performed to remove redundant bounding boxes.
The standard procedure is to identify boxes with an intersection-over-union (IoU) greater than a fixed threshold and remove the box with the lower class probability.
This strategy is inappropriate for our method for two reasons.
First, it does not consider the predicted variance of the bounding boxes.
Second, the class probability in our case does not indicate the quality of the bounding box.
For example, it is relatively easy to classify the front of a semi-truck as a vehicle, but it is difficult to accurately predict its extents.

Alternatively, we propose an adaptive NMS algorithm, which uses the predicted variance to determine an appropriate IoU threshold for a pair of boxes.
In addition, we utilize the likelihood of the box as its score.
Since we use the means of the mixture model, the likelihood of the box corresponding to the $k$th component of the mixture model reduces to $\alpha_k / 2\hat{\sigma}_{k}$.

\begin{figure}
\centering
\includegraphics[width=0.45\textwidth]{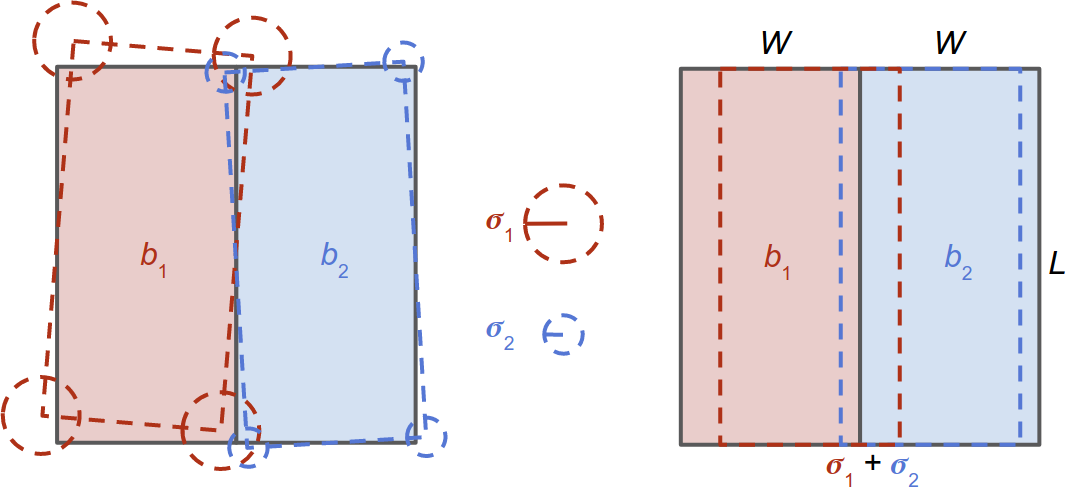}
\caption{An illustration of our adaptive NMS technique. Consider a pair of vehicles positioned side-by-side. The dashed outlines on the left depict a possible set of predictions produced by our method. To determine whether the bounding boxes encapsulate unique objects, we utilize the predicted variances (shown in the center) to estimate the worst-case overlap as shown on the right. In this example, both bounding boxes would be preserved because the actual overlap is less than the estimated worst-case overlap.}
\label{fig:adaptive-nms}
\end{figure}

In the top-down view, bounding boxes should not overlap; however, due to the uncertainty in the predictions, some amount of overlap is expected.
For each pair of overlapping bounding boxes from the same class, we calculate the upper-bound on the IoU given their predicted variances.
We accomplish this by assuming the worst-case scenario, which occurs when two objects of the same dimensions are side-by-side, as shown in Figure \ref{fig:adaptive-nms}.
In this case, the maximum tolerated IoU $t$ should be,
\begin{equation}
t =
\begin{cases}
      \frac{\sigma_1 + \sigma_2}{2w - \sigma_1 - \sigma_2} & \sigma_1 + \sigma_2 < w \\
      \hfil 1 & \text{otherwise} \\
 \end{cases}
\end{equation}
where $w$ is the average width of the object's bounding box. For example, $w\approx2$ meters for vehicles.

If the IoU of a pair of boxes exceeds the maximum threshold, then either one of the predicted boxes is incorrect or one of the predicted variances is incorrect.
Assuming the bounding box is wrong, we can remove the box with the lower likelihood.
Otherwise, we can assume the variance is inaccurate and increase the variance of the less likely box such that $t$ would equal the observed IoU.
The former is referred to as Hard NMS while the latter is analogous to Soft NMS \cite{softnms}. The effect of this choice is examined in the ablation study in Section~\ref{sec:ablation}.

\section{Experiments}
\label{sec:experiments}
Our proposed approach is evaluated and compared against state-of-the-art methods on two datasets: the large-scale ATG4D object detection dataset, and the small-scale KITTI object detection benchmark \cite{Geiger2012CVPR}.

\subsection{Evaluation on ATG4D}
\label{sec:eval_car3d}
The ATG4D dataset contains 5,000 sequences for training and 500 for validation. The sequences from the train set are sampled at $10$ Hz, while the validation set are sampled at $0.5$ Hz. The entire dataset contains 1.2 million sweeps for training, and 5,969 sweeps for validation. All sweeps are captured using a Velodyne 64E LiDAR. We adopt the same evaluation procedure as the KITTI benchmark and only consider the detections within the front $90^\circ$ field of view of the sensor and up to $70$ meters.
As a result, our range image contains only this front portion of the LiDAR data and has dimensions of $512 \times 64$.

The network is trained for 300k iterations using the Adam optimizer \cite{adam} with a learning rate of 0.002 exponential decayed at a rate of 0.99 every 150 iterations. We use a batch size of 128 distributed over 32 GPUs. For vehicles, we learn a multimodal distribution with three components ($K=3$), and for pedestrians and bikes, we learn a unimodal distribution ($K=1$).

Table \ref{tab:car3d_detection_results} shows our results on the validation set compared to other state-of-the-art methods.
Like the KITTI benchmark, we compute the average precision (AP) at a $0.7$ IoU for vehicles and a $0.5$ IoU for bikes and pedestrians.
On this dataset, our approach out-performs existing state-of-the-art methods in the $0-70$ meter range.
Furthermore, our method out-performs the LiDAR-only methods at all ranges, and it is only surpassed by a LiDAR+RGB method on vehicles and bikes at long range where the additional image data provides the most value. We believe our approach performs significantly better on pedestrians because our range view representation does not discretize the input data into voxels.

\begin{table*}[t]
\centering
\caption{BEV Object Detection Performance on ATG4D}
\vspace{-0.75em}
\scalebox{0.8}{
    \begin{tabular}{c|c|cccc|cccc|cccc}
    \hline
    \multirow{2}{*}{Method} & \multirow{2}{*}{Input} & \multicolumn{4}{c|}{Vehicle $AP_{0.7}$ }& \multicolumn{4}{c|}{Bike $AP_{0.5}$} & \multicolumn{4}{c}{Pedestrian $AP_{0.5}$}\\
    & & 0-70m & 0-30m & 30-50m & 50-70m & 0-70m & 0-30m & 30-50m & 50-70m & 0-70m & 0-30m & 30-50m & 50-70m\\
    \hline
    LaserNet (Ours) & LiDAR  & \textbf{85.34} & \textbf{95.02} & \textbf{84.42} & 67.65  & \textbf{61.93} & \textbf{74.62} & 51.37 & 40.95  & \textbf{80.37} & \textbf{88.02} & \textbf{77.85} & \textbf{65.75}\\
    PIXOR ~\cite{pixor} & LiDAR  & 80.99  & 93.34  & 80.20 & 60.19  & - & - & -  & - & - & - & - & -\\
    PIXOR++ ~\cite{hdnet} & LiDAR  & 82.63  & 93.80 & 82.34  & 63.42  & - & - & -  & - & - & - & - & -\\
    ContFuse ~\cite{ibm} & LiDAR  & 83.13  & 93.08  & 82.48  & 65.53  & 57.27  & 68.08  & 48.83 & 38.26  & 73.51  & 80.60 & 71.68  & 59.12\\
    \hline
    ContFuse ~\cite{ibm} & LiDAR+RGB  & 85.17  & 93.86  & 84.41  & \textbf{69.83} & 61.13  & 72.01  & \textbf{52.60} & \textbf{43.03} & 76.84 & 82.97 & 75.54 & 64.19\\
    \hline
    \end{tabular}
}
\label{tab:car3d_detection_results}
\vspace{-0.25em}
\end{table*}

\subsection{Ablation Study on ATG4D}
\label{sec:ablation}

\begin{table*}[t]
\centering
\caption{Ablation Study on ATG4D}
\vspace{-0.75em}
\scalebox{0.8}{
  \begin{tabular}{ccccc|c}
    \hline
    Predicted Distribution & Image Spacing & Mean Shift & IoU Threshold & NMS Type & Vehicle $AP_{0.7}$\\
    \hline
    Mean-only  & Laser   & Yes & 0.1      & Hard & 77.05\\
    \hline
    Unimodal   & Uniform & Yes & 0.1      & Hard & 79.14\\
    Unimodal   & Laser   & No  & 0.1      & Hard & 80.22\\
    Unimodal   & Laser   & Yes & 0.1      & Hard & 80.92\\
    \hline
    Multimodal & Laser   & Yes & 0.1      & Hard & 81.80\\
    Multimodal & Laser   & Yes & N/A      & Soft & 84.43\\
    Multimodal & Laser   & Yes & Adaptive & Hard & 83.68\\
    Multimodal & Laser   & Yes & Adaptive & Soft & \textbf{85.34}\\
    \hline
  \end{tabular}
}
\label{tab:car3d_ablation_study}
\vspace{-0.75em}
\end{table*}

In this section, we examine the various aspects of our proposed method, and their effects on vehicle detection performance.
We conduct the ablation study on the ATG4D dataset, and the results are shown in Table~\ref{tab:car3d_ablation_study}.
A detailed description of the study is provided below.

\medskip\noindent\textit{Predicting a Probability Distribution}\\
\noindent The largest improvement is a result of predicting a distribution of bounding boxes instead of merely the mean.
When only predicting the mean bounding box, equation (\ref{eqn:cluster-corners}) becomes a simple average, and equation (\ref{eqn:cluster-loss}) reduces to the $\ell_1$ loss on the box corners.
Furthermore, the score of the bounding box is the class probability in this case.
We believe the loss in performance is due to the probability not being well correlated with the accuracy of the box.

\medskip\noindent\textit{Image Formation}\\
\noindent Previous methods which utilize the range view \cite{mv3d, velofcn} uniformly discretize the elevation angle into rows.
However, the lasers in the Velodyne 64E LiDAR are not uniformly spaced.
A gain in performance can be obtained by mapping points to rows using the laser id and processing the data directly as the sensor captured it.

\medskip\noindent\textit{Mean Shift Clustering}\\
\noindent Without clustering, each point independently predicts a distribution of bounding boxes.
These individual box predictions naturally contain some amount of noise, and we can reduce this noise by combining the independent predictions through mean shift clustering.

\medskip\noindent\textit{Non-Maximum Suppression}\\
\noindent When the LiDAR points are sparse, there are multiple configurations of bounding boxes that could explain the observed data.
By predicting a multimodal distribution at each point, we can further improve the recall of our method.
It is inappropriate to use NMS with a strict threshold when producing a multimodal distribution because only the most likely bounding box will persist.
Alternatively, we could use Soft NMS \cite{softnms} to re-weight the confidence instead of eliminating boxes, but this breaks our probabilistic interpretation of the confidence.
By leveraging our adaptive NMS algorithm, we maintain our probabilistic interpretation and obtain better performance.

\begin{table}[t]
\centering
\caption{Runtime Performance on KITTI}
\vspace{-0.75em}
\scalebox{0.8}{
  \begin{tabular}{ lcc }
    \hline
    Method &  Forward Pass (ms) & Total (ms) \\
    \hline
    LaserNet (Ours) & \textbf{12} & \textbf{30} \\
    PIXOR \cite{pixor} & 35 & 62 \\
    PIXOR++ \cite{hdnet} & 35 & 62 \\
    VoxelNet \cite{voxelnet} & 190 & 225 \\
    \hline
    MV3D \cite{voxelnet} & - & 360 \\
    AVOD \cite{avod} & 80 & 100 \\
    F-PointNet \cite{frustum_pointnet} & - & 170 \\
    ContFuse \cite{ibm} & 60 & - \\
    \hline
  \end{tabular}
}
\label{tab:run_time_results}
\end{table}

\subsection{Runtime Evaluation}
Runtime performance is equally important for the purpose of autonomous driving.
Table \ref{tab:run_time_results} compares the runtime performance between our approach (measured on a NVIDIA 1080Ti GPU) and existing methods on KITTI.
The forward pass refers to the amount of time it takes to run the network, and the total includes pre and post processing in addition to the forward pass.
Our proposed method is twice as fast as the fastest state-of-the-art method.
We are able to achieve a faster runtime because we operate on a small dense range view image instead of a large and sparse bird's eye view representation.

\begin{table}[t]
\centering
\caption{BEV Object Detection Performance on KITTI}
\vspace{-0.75em}
\scalebox{0.8}{
  \begin{tabular}{ lcccc }
    \hline
    \multirow{2}{*}{Method} &  \multirow{2}{*}{Input}  & \multicolumn{3}{c}{Vehicle $AP_{0.7}$} \\
     & & Easy & Moderate & Hard\\
    \hline
    LaserNet (Ours) & LiDAR & 78.25 & 73.77 & 66.47 \\
    PIXOR \cite{pixor} & LiDAR & 81.70 & 77.05 & 72.95 \\
    PIXOR++ \cite{hdnet} & LiDAR & \textbf{89.38} & 83.70 & \textbf{77.97} \\
    VoxelNet \cite{voxelnet} & LiDAR & 89.35 & 79.26 & 77.39 \\
    \hline
    MV3D \cite{mv3d} & LiDAR+RGB & 86.02 & 76.90 & 68.49 \\
    AVOD \cite{avod} & LiDAR+RGB & 88.53 & 83.79 & 77.90 \\
    F-PointNet \cite{frustum_pointnet} & LiDAR+RGB & 88.70 & 84.00 & 75.33 \\
    ContFuse \cite{ibm} & LiDAR+RGB & 88.81 & \textbf{85.83} & 77.33 \\
    \hline
  \end{tabular}
}
\label{tab:kitti_detection_results}
\end{table}

\begin{figure*}
\begin{subfigure}{.5\textwidth}
  \centering
  \includegraphics[width=.85\linewidth]{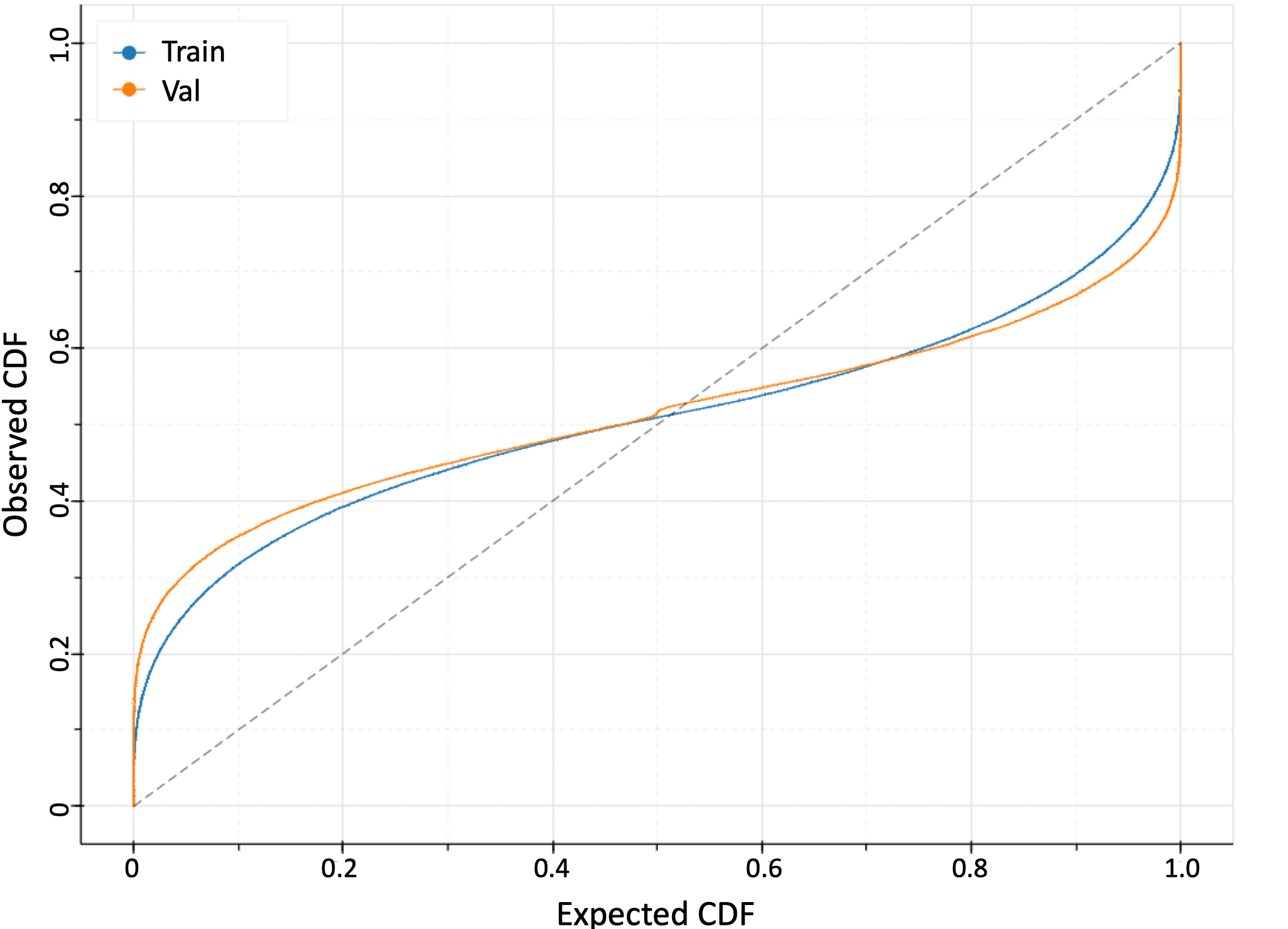}
  \caption{Calibration on KITTI}
  \label{fig:variance_kitti}
\end{subfigure}\vspace{2mm}
\begin{subfigure}{.5\textwidth}
  \centering
  \includegraphics[width=.85\linewidth]{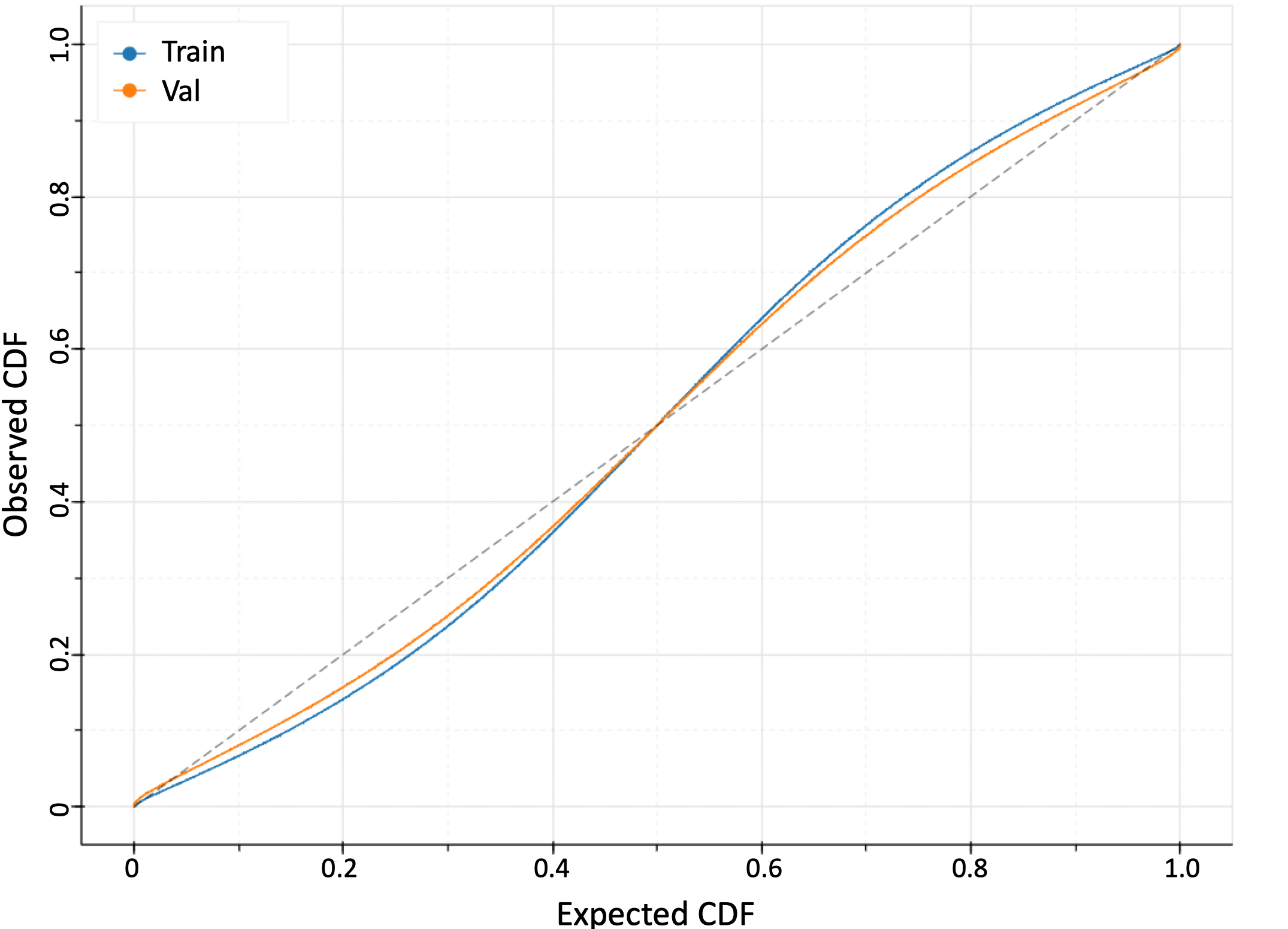}
  \caption{Calibration on ATG4D}
  \label{fig:variance_car3d}
\end{subfigure}
\caption{Plots showing the calibration of the predicted distribution over bounding boxes on the train and validation sets. A perfectly calibrated distribution corresponds to a line with unit slope (dashed line in the plots). We observe that the model is unable to learn the probability distribution on KITTI, whereas it is capable of learning the distribution on the larger ATG4D.}
\label{fig:variance}
\end{figure*}

\subsection{Evaluation on KITTI}
The KITTI object detection benchmark \cite{Geiger2012CVPR} contains 7,481 training sweeps and 7,518 testing sweeps captured by a Velodyne 64E LiDAR.
For the training set, object annotations are provided within the front $90^\circ$ field of view of the LiDAR up to approximately $70$ meters.

To train our network, we use 5,985 sweeps from the training set, and we holdout the remaining sweeps for validation.
We train the network for 50k iterations with the same learning schedule as before, and we use a batch size of 12 on a single GPU.
To help avoid overfitting on this small training set, we use data augmentation.
We randomly flip the range image, and we apply random pixel shifts in the horizontal dimension.

Learning a probability distribution, especially a multimodal distribution, over bounding boxes proved to be difficult on such a small dataset.
Therefore, we trained the network to only detect vehicles and to predict a unimodal probability distribution over bounding boxes.

The KITTI evaluation server computes the AP at a $0.7$ IoU across three difficulty levels: easy, moderate, and hard.
As shown in Table~\ref{tab:kitti_detection_results}, our approach performs worse than current state-of-the-art bird's eye view detectors on this small dataset. The following section explores this discrepancy in performance between the results on the large dataset shown in Table~\ref{tab:car3d_detection_results} and the results on the small KITTI dataset.

\subsection{Analysis of the Predicted Distribution}
\label{sec:analysis}
On the small dataset, our approach under-performs compared to state-of-the-art methods.
While on a significantly larger dataset, our method out-performs the previous work.
The ablation study in Section~\ref{sec:ablation} emphasizes the importance of estimating a probability distribution to our approach.
If the network is unable to accurately learn this distribution, our method will perform sub-optimally.

To evaluate the quality of the predicted distributions of a learned model, we compute the probability of each ground truth label given our predicted probability distribution, and we plot the expected CDF versus the observed CDF as shown in Figure~\ref{fig:variance}.
We perform this evaluation on both the KITTI and the ATG4D datasets.
On both datasets, the model is trained to predict a unimodal probability distribution.
With the small scale KITTI dataset, our model is unable to correctly estimate the probability distribution.
However, the model is capable of precisely learning the distribution on the large scale ATG4D dataset.
We hypothesize that learning the distribution requires the network to see many more examples than are available in the KITTI training set, and this helps explain the difference in model performance across these two datasets.

\section{Discussion}

In recent years, the research community has generally favored detectors that operate in the bird's eye view or directly on the 3D point cloud. There are many advantages to these approaches. Operating directly in the output space allows a detector to have a consistent prior over the object's shape, which can make the learning problem easier. This is especially important when training on smaller datasets.

However, we think it is a mistake to overlook the range view representation. Deep learning approaches have consistently shown success across many domains when applied to the raw input data without hand-engineered feature extraction, projections, or other manipulations given there is enough training data. The range view is the native representation of the LiDAR data; as a result, it is naturally compact. This compact representation leads to significant gains in efficiency. Moreover, it inherently conveys important contextual information in terms of how the data was captured, and this context can be lost when the range data is projected into a point cloud. At the same time, the range view poses significant challenges to learning due to the varying scale and shape of objects as well as the need to handle occlusions. With a large training set, we have shown that it is possible to overcome these challenges and produce competitive results while being more computationally efficient.

On a smaller dataset, our approach does not achieve the same performance as the state-of-the-art bird's eye view detectors. Key elements of our approach include operating in the native view of the sensor and predicting a probability distribution over bounding boxes. Both of these make the learning problem more difficult and require more training data to perform well.

Finally, although we have focused on one specific representation in this paper, we believe that other detection approaches would also benefit from predicting probability distributions over bounding boxes.



{\small
\bibliographystyle{ieee_fullname}
\bibliography{egbib}

\begin{thebibliography}{10}\itemsep=-1pt

\bibitem{birdnet}
Jorge Beltr{\'a}n, Carlos Guindel, Francisco~Miguel Moreno, Daniel Cruzado,
  Fernando Garc{\'\i}a, and Arturo De~La~Escalera.
\newblock Birdnet: A 3{D} object detection framework from {LiDAR} information.
\newblock In {\em Proceedings of the International Conference on Intelligent
  Transportation Systems (ITSC)}, 2018.

\bibitem{softnms}
Navaneeth Bodla, Bharat Singh, Rama Chellappa, and Larry~S Davis.
\newblock Soft-{NMS} -- {I}mproving object detection with one line of code.
\newblock In {\em Proceedings of the IEEE International Conference on Computer
  Vision (ICCV)}, 2017.

\bibitem{cvpr16chen}
Xiaozhi Chen, Kaustav Kundu, Ziyu Zhang, Huimin Ma, Sanja Fidler, and Raquel
  Urtasun.
\newblock Monocular 3{D} object detection for autonomous driving.
\newblock In {\em Proceedings of the IEEE Conference on Computer Vision and
  Pattern Recognition (CVPR)}, 2016.

\bibitem{nips15chen}
Xiaozhi Chen, Kaustav Kundu, Yukun Zhu, Andrew~G Berneshawi, Huimin Ma, Sanja
  Fidler, and Raquel Urtasun.
\newblock 3{D} object proposals for accurate object class detection.
\newblock In {\em Proceedings of Advances in Neural Information Processing
  Systems (NIPS)}, 2015.

\bibitem{mv3d}
Xiaozhi Chen, Huimin Ma, Ji Wan, Bo Li, and Tian Xia.
\newblock Multi-view 3{D} object detection network for autonomous driving.
\newblock In {\em Proceedings of the IEEE Conference on Computer Vision and
  Pattern Recognition (CVPR)}, 2017.

\bibitem{uncertainty_3d_detection_itsc}
Di Feng, Lars Rosenbaum, and Klaus Dietmayer.
\newblock Towards safe autonomous driving: Capture uncertainty in the deep
  neural network for lidar 3{D} vehicle detection.
\newblock In {\em Proceedings of the International Conference on Intelligent
  Transportation Systems (ITSC)}, 2018.

\bibitem{uncertainty_3d_detection}
Di Feng, Lars Rosenbaum, Fabian Timm, and Klaus Dietmayer.
\newblock Leveraging heteroscedastic aleatoric uncertainties for robust
  real-time {LiDAR} 3{D} object detection.
\newblock {\em arXiv preprint arXiv:1809.05590}, 2018.

\bibitem{Geiger2012CVPR}
Andreas Geiger, Philip Lenz, and Raquel Urtasun.
\newblock Are we ready for autonomous driving? the {KITTI} vision benchmark
  suite.
\newblock In {\em Proceedings of the IEEE Conference on Computer Vision and
  Pattern Recognition (CVPR)}, 2012.

\bibitem{rgb_dense_lidar_multiview_fusion}
Alejandro Gonz{\'a}lez, Gabriel Villalonga, Jiaolong Xu, David V{\'a}zquez,
  Jaume Amores, and Antonio~M L{\'o}pez.
\newblock Multiview random forest of local experts combining {RGB} and {LIDAR}
  data for pedestrian detection.
\newblock In {\em Proceedings of the IEEE Intelligent Vehicles Symposium (IV)},
  2015.

\bibitem{multiple-choice}
Abner Guzman-Rivera, Dhruv Batra, and Pushmeet Kohli.
\newblock Multiple choice learning: Learning to produce multiple structured
  outputs.
\newblock In {\em Proceedings of Advances in Neural Information Processing
  Systems (NIPS)}, 2012.

\bibitem{resnet}
Kaiming He, Xiangyu Zhang, Shaoqing Ren, and Jian Sun.
\newblock Deep residual learning for image recognition.
\newblock In {\em Proceedings of the IEEE Conference on Computer Vision and
  Pattern Recognition (CVPR)}, 2016.

\bibitem{uncertainty_2d_detection}
Borui Jiang, Ruixuan Luo, Jiayuan Mao, Tete Xiao, and Yuning Jiang.
\newblock Acquisition of localization confidence for accurate object detection.
\newblock In {\em Proceedings of the European Conference on Computer Vision
  (ECCV)}, 2018.

\bibitem{kendalluncertainities}
Alex Kendall and Yarin Gal.
\newblock What uncertainties do we need in bayesian deep learning for computer
  vision?
\newblock In {\em Proceedings of Advances in Neural Information Processing
  Systems (NIPS)}, 2017.

\bibitem{adam}
Diederik~P Kingma and Jimmy Ba.
\newblock Adam: A method for stochastic optimization.
\newblock {\em arXiv preprint arXiv:1412.6980}, 2014.

\bibitem{avod}
Jason Ku, Melissa Mozifian, Jungwook Lee, Ali Harakeh, and Steven~L Waslander.
\newblock Joint 3{D} proposal generation and object detection from view
  aggregation.
\newblock In {\em Proceedings of the IEEE/RSJ International Conference on
  Intelligent Robots and Systems (IROS)}, 2018.

\bibitem{velofcn}
Bo Li, Tianlei Zhang, and Tian Xia.
\newblock Vehicle detection from 3{D} lidar using fully convolutional network.
\newblock In {\em Proceedings of Robotics: Science and Systems (RSS)}, 2016.

\bibitem{ibm}
Ming Liang, Bin Yang, Shenlong Wang, and Raquel Urtasun.
\newblock Deep continuous fusion for multi-sensor 3{D} object detection.
\newblock In {\em Proceedings of the European Conference on Computer Vision
  (ECCV)}, 2018.

\bibitem{retinanet}
Tsung-Yi Lin, Priya Goyal, Ross Girshick, Kaiming He, and Piotr Doll{\'a}r.
\newblock Focal loss for dense object detection.
\newblock In {\em Proceedings of the IEEE International Conference on Computer
  Vision (ICCV)}, 2017.

\bibitem{ssd}
Wei Liu, Dragomir Anguelov, Dumitru Erhan, Christian Szegedy, Scott Reed,
  Cheng-Yang Fu, and Alexander~C Berg.
\newblock {SSD}: Single shot multibox detector.
\newblock In {\em Proceedings of the European Conference on Computer Vision
  (ECCV)}, 2016.

\bibitem{cvpr17mousavian}
Arsalan Mousavian, Dragomir Anguelov, John Flynn, and Jana Kosecka.
\newblock 3{D} bounding box estimation using deep learning and geometry.
\newblock In {\em Proceedings of the IEEE Conference on Computer Vision and
  Pattern Recognition (CVPR)}, 2017.

\bibitem{rgb_dense_lidar_fusion}
Cristiano Premebida, Jo{\~{a}}o Carreira, Jorge Batista, and Urbano Nunes.
\newblock Pedestrian detection combining {RGB} and dense {LIDAR} data.
\newblock In {\em Proceedings of the IEEE/RSJ International Conference on
  Intelligent Robots and Systems (IROS)}, 2014.

\bibitem{frustum_pointnet}
Charles~R Qi, Wei Liu, Chenxia Wu, Hao Su, and Leonidas~J Guibas.
\newblock Frustum pointnets for 3{D} object detection from {RGB-D} data.
\newblock In {\em Proceedings of the IEEE Conference on Computer Vision and
  Pattern Recognition (CVPR)}, 2018.

\bibitem{pointnet}
Charles~R Qi, Hao Su, Kaichun Mo, and Leonidas~J Guibas.
\newblock Pointnet: Deep learning on point sets for 3{D} classification and
  segmentation.
\newblock In {\em Proceedings of the IEEE Conference on Computer Vision and
  Pattern Recognition (CVPR)}, 2017.

\bibitem{faster_rcnn}
Shaoqing Ren, Kaiming He, Ross Girshick, and Jian Sun.
\newblock Faster {R-CNN}: Towards real-time object detection with region
  proposal networks.
\newblock In {\em Proceedings of Advances in Neural Information Processing
  Systems (NIPS)}, 2015.

\bibitem{cvpr18xu}
Bin Xu and Zhenzhong Chen.
\newblock Multi-level fusion based 3{D} object detection from monocular images.
\newblock In {\em Proceedings of the IEEE Conference on Computer Vision and
  Pattern Recognition (CVPR)}, 2018.

\bibitem{pointfusion}
Danfei Xu, Dragomir Anguelov, and Ashesh Jain.
\newblock Pointfusion: Deep sensor fusion for 3{D} bounding box estimation.
\newblock In {\em Proceedings of the IEEE Conference on Computer Vision and
  Pattern Recognition (CVPR)}, 2018.

\bibitem{hdnet}
Bin Yang, Ming Liang, and Raquel Urtasun.
\newblock {HDNET}: Exploiting {HD} maps for 3d object detection.
\newblock In {\em Proceedings of the Conference on Robot Learning (CoRL)},
  2018.

\bibitem{pixor}
Bin Yang, Wenjie Luo, and Raquel Urtasun.
\newblock {PIXOR}: Real-time 3{D} object detection from point clouds.
\newblock In {\em Proceedings of the IEEE Conference on Computer Vision and
  Pattern Recognition (CVPR)}, 2018.

\bibitem{dla}
Fisher Yu, Dequan Wang, Evan Shelhamer, and Trevor Darrell.
\newblock Deep layer aggregation.
\newblock In {\em Proceedings of the IEEE Conference on Computer Vision and
  Pattern Recognition (CVPR)}, 2018.

\bibitem{voxelnet}
Yin Zhou and Oncel Tuzel.
\newblock Voxelnet: End-to-end learning for point cloud based 3{D} object
  detection.
\newblock In {\em Proceedings of the IEEE Conference on Computer Vision and
  Pattern Recognition (CVPR)}, 2018.

\end{thebibliography}
}
\end{document}